\documentclass[letterpaper]{article} 
\usepackage{aaai2027}  
\nocopyright

\usepackage[hyphens]{url}  
\usepackage{graphicx} 
\usepackage{natbib}  
\usepackage{caption} 
\usepackage{algorithm}
\usepackage{algorithmic}
\usepackage{booktabs}
\usepackage{amsmath,amssymb}
\usepackage{pifont}
\usepackage{tikz}
\usepackage{multirow}
\usepackage{threeparttable}
\usepackage[most]{tcolorbox}
\usepackage[table]{xcolor}
\usepackage{tabularx}

\usetikzlibrary{arrows.meta, positioning, fit, backgrounds}
\newcommand{\DACCI}{\textsc{D\textsuperscript{2}ACCI}}
\newcommand{\MiMemoryStack}{\textsc{MemStack}}
\newcommand{\locomo}{\textsc{LoCoMo}}
\newcommand{\lme}{\textsc{LongMemEval}}
\newcommand{\pmem}{\textsc{PersonaMem-V2}}
\newcommand{\DCov}{\textsc{DCR}}

\title{\DACCI{}: A Dual-Loop Diagnostic Protocol for Evidence-Preserving Agent Memory}
\author{
    Xule Liu,
    Yijun Liu,
    Chao Li,
    Shao Kun\thanks{Corresponding author.}
}
\affiliations{
    \textsuperscript{\rm}Xiaomi Inc.\\
    liuxule@xiaomi.com, liuyijun3@xiaomi.com, \\ lichao75@xiaomi.com, shaokun@xiaomi.com
}
\begin{document}
\maketitle

\begin{abstract}

Memory is a key capability of LLM agents.
Persistent memory extends this across sessions---enabling recall, revision, and personalization.
Yet its multi-stage pipeline (ingestion, retrieval, filtering, generation) makes failures difficult to localize: end-to-end evaluation reveals that an error occurred, but not which stage caused it.
Existing evaluations often report aggregate performance without paired statistical comparisons, slice-level non-regression checks, or stage-level diagnostic traces.
We propose \textbf{\DACCI{}} (\textbf{D}iagnostic-\textbf{D}riven \textbf{A}rtifact-based \textbf{C}losed-loop \textbf{C}ontrolled \textbf{I}teration), a dual-loop protocol whose outer diagnostic gate promotes, feature-flags, or rejects memory interventions based on paired evidence, protected-slice monitoring, and trace-level localizability.
We further introduce \DCov{}, a graded observability metric that measures whether failures remain localizable, and \DACCI{}-Eval, a reusable artifact for gate replay.
We instantiate the protocol in \MiMemoryStack{} and evaluate on three public benchmarks, achieving 93.59\% on \locomo{}, 90.93\% on \lme{}, and 57.20\% on \pmem{}.
Five paired ablations show that supplement extraction, session-memory retrieval, and Forget Guard yield statistically significant gains (+1.9 to +3.7pp, all $p \le .003$).
In contrast, BM25/RRF is retained as a monitored feature flag---a distinction invisible to aggregate-only evaluation.
A diagnostic audit shows enriched traces substantially improve root-cause agreement over result-only relabeling.
Diagnostic artifacts reach 98--100\% \DCov{}@3 versus 0\% for results-only logs.
These results establish that robust memory-system iteration demands traceable, statistically grounded, and regression-aware evidence---exactly the gap \DACCI{} fills.

\end{abstract}

\section{Introduction}
LLM agents are increasingly expected to operate over long time horizons, where the information needed for a later interaction may no longer fit within a limited context window.
Persistent memory addresses this limitation by storing, organizing, and reusing information across interactions, and is becoming a first-class component of LLM agents~\cite{packer2023memgpt,chhikara2025mem0,evermemos2026acl}.  
A deployed assistant must recall facts across months of interaction, distinguish stable preferences from transient context, respect revised information, and avoid using memories that users ask to forget~\cite{maharana2024locomo,wu2025longmemeval,jiang2025personamem}.  
These requirements are difficult to satisfy reliably because persistent-memory systems involve multiple processing stages. 
When a memory-augmented answer is incorrect, the reason may lie in failed extraction, incorrect consolidation with another memory, irrelevant retrieval, over-aggressive filtering, or omission of a critical update from the final prompt.
End-to-end evaluation reveals that an error occurred, but not which stage caused it or what improvements should be applied.

\begin{figure}[t]
\centering
\includegraphics[width=\columnwidth]{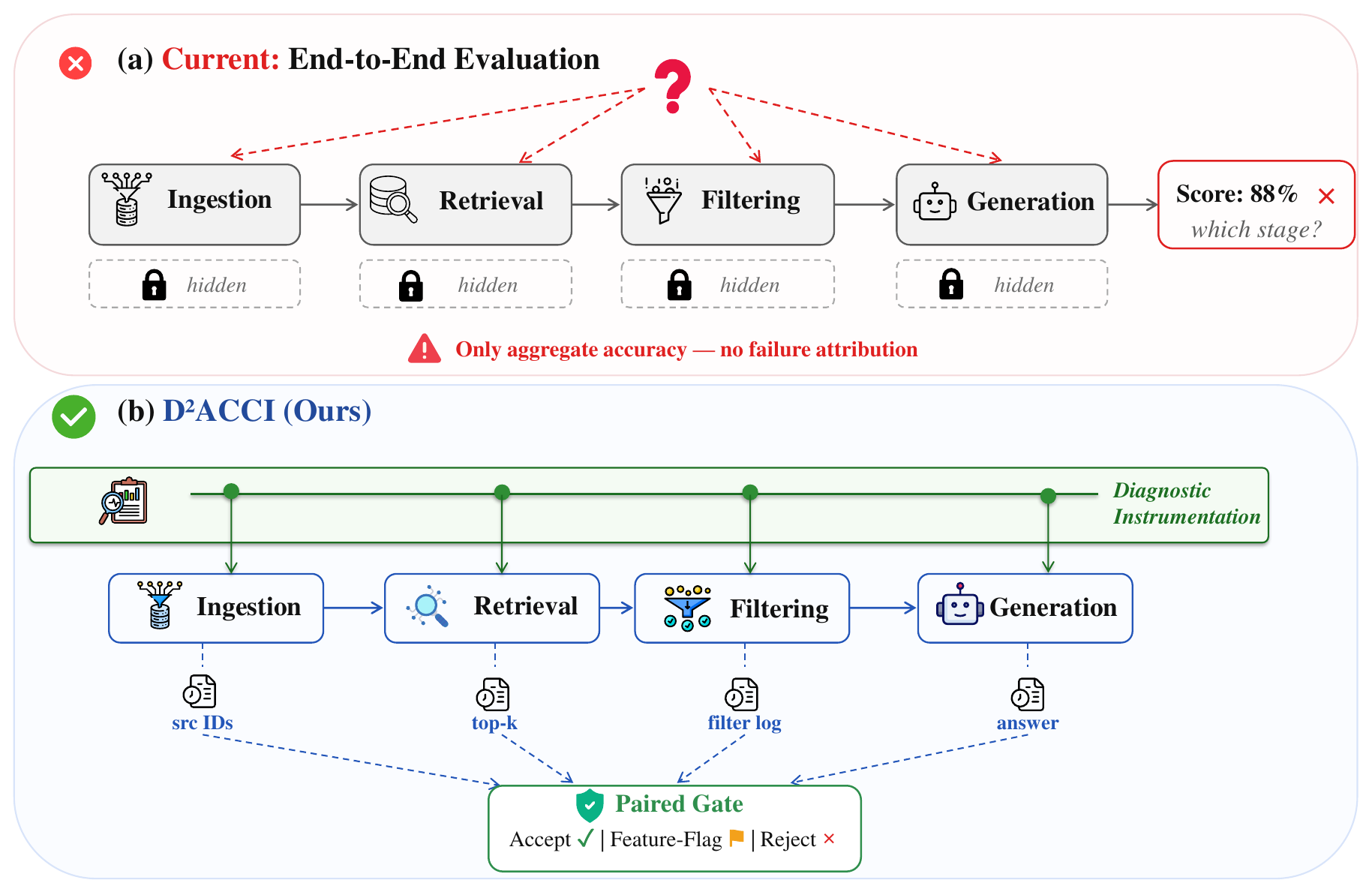}
\caption{(a)~Current end-to-end evaluation reports only aggregate accuracy; internal pipeline states are hidden and failures cannot be attributed to a specific stage. (b)~\DACCI{} emits per-stage traces, enabling paired gates to localize failures and drive targeted repairs.}
\label{fig:motivation}
\end{figure}

We therefore argue that agent memory should be evaluated as an \emph{evolvable runtime system}.  
A long-lived agent is repeatedly updated as developers revise retrieval policies, consolidation rules, deduplication thresholds, and memory management logics~\cite{gao2025selfevolving}. 
Without explicit regression controls, these updates can silently degrade the user experience or performance of task slices, such as temporal inference or preference fidelity.  
This risk is especially pronounced for personalized memory because improvements are often non-monotonic: broader retrieval may recover temporally relevant evidence while introducing conflicting preferences, whereas stronger deduplication may reduce redundancy while deleting rare but crucial facts.

For agent builders, this makes memory iteration a deployment problem rather than only a benchmark problem. 
Ideal iteration requires more than aggregate component ablations.
Prior evaluations of agent memory rarely combine paired comparisons, protected slice non-regression checks, null result preservation, and stage-level diagnostics~\cite{chhikara2025mem0,evermemos2026acl,rasmussen2025zep}.
Thus, an aggregate improvement may obscure trade offs or regressions under specific benchmarks.
Reliable memory system evolution requires a reproducible procedure that links each intervention to a testable failure hypothesis, statistically supported paired evidence, protected slice monitoring, and traces capable of preserving stage-level failure localization.

To address this gap, we introduce \textbf{\DACCI{}} (\textbf{D}iagnostic-\textbf{D}riven \textbf{A}rtifact-based \textbf{C}losed-loop \textbf{C}ontrolled \textbf{I}teration), a dual-loop protocol that separates an inner runtime loop from an outer diagnostic-evolution loop. 
The inner loop executes the memory-augmented agent, whereas the outer loop evaluates candidate changes and determines whether they should be promoted, feature flagged, or rejected. 
D\textsuperscript{2}ACCI defines a decision contract that ties these decisions to paired statistical evidence, protected slice non-regression checks, and trace-level localizability.

Additionally, we introduce \DCov{}, a graded trace-coverage metric for measuring whether failures remain localizable at the relevant stage, and \DACCI{}-Eval, a reusable evaluation artifact supporting paired statistical analysis, protected-slice gates, discordance export, and deterministic gate replay. 
The artifact preserves both positive and null findings, allowing later iterations to reuse prior evidence.

We instantiate D\textsuperscript{2}ACCI in \MiMemoryStack{}, a diagnosable memory kernel with multi-granularity storage and feature-flagged retrieval.
We evaluate on three public benchmarks, achieving 93.59\% on \locomo{}, 90.93\% on \lme{}, and 57.20\% on \pmem{}.
Across five paired feature-gated configurations, supplement extraction, session-memory retrieval, and Forget Guard yield statistically significant gains of +2.71, +3.67, and +1.92 percentage points, respectively, whereas BM25/RRF yields no statistically significant improvement.
Diagnostic artifacts achieve 98--100\% \DCov{}@3, compared with 0\% for results-only logs.
On an external Mem0-compatible slice, trace-adapted and results-only outputs achieve the same task accuracy (43.33\%) but differ substantially in diagnostic coverage, demonstrating artifact portability beyond task performance.

Our contributions are as follows:
\begin{enumerate}
\setlength{\itemsep}{1pt}
\setlength{\parskip}{0pt}

\item \textbf{\DACCI{} protocol and decision contract.}
We introduce a dual-loop diagnostic protocol that promotes, feature-flags, or rejects memory interventions based on paired statistical evidence, protected-slice non-regression checks, and stage-level trace sufficiency.  The individual primitives (paired runs, traces, feature flags) are familiar; the contribution is the decision contract that binds them into acceptance criteria---changing outcomes that aggregate-only reasoning cannot (e.g., issuing contradictory BM25/RRF verdicts across benchmarks, or promoting components without slice monitors).

\item \textbf{\DACCI{}-Eval artifact and \DCov{} metric.}
We provide a reusable evaluation artifact for paired comparisons, protected slice gates, discordance export, and deterministic gate replay, together with a graded metric for measuring failure localizability.

\item \textbf{\MiMemoryStack{} case study.}
We instantiate D\textsuperscript{2}ACCI in a diagnosable memory kernel with multi-granularity storage and feature-gated retrieval, and evaluate it on three public benchmarks with distinct failure profiles.  A separate Mem0-compatible slice demonstrates artifact portability beyond the primary system.
\end{enumerate}

\begin{algorithm}[t]
\caption{\DACCI{} dual-loop memory-system iteration}
\label{alg:dacci-iteration}
\begin{algorithmic}[1]
\STATE \textbf{Inner loop:} run $\mathcal{P}_{\theta}$ to ingest, update, retrieve, constrain, assemble, and answer while emitting $d_i$.
\STATE Specify outer-loop hypothesis $H_t$, target slice, and expected failure mechanism.
\STATE Implement candidate $\theta'{}$ in the shared core or behind a feature flag.
\STATE Run baseline/candidate and collect details JSONL plus diagnostics.
\STATE Compute paired outcomes: $I$, $R$, $B_w$, $B_c$.
\STATE Label root causes in $R \cup B_w$ using the evidence-loss ladder.
\STATE Run paired gate: McNemar/bootstrap statistics, slice checks, and \DCov{}.
\STATE \textbf{Outer loop:} accept, rollback, or gate the feature; archive artifacts and rejected priors.
\end{algorithmic}
\end{algorithm}

\begin{figure*}[t]
\centering
\includegraphics[width=0.75\textwidth]{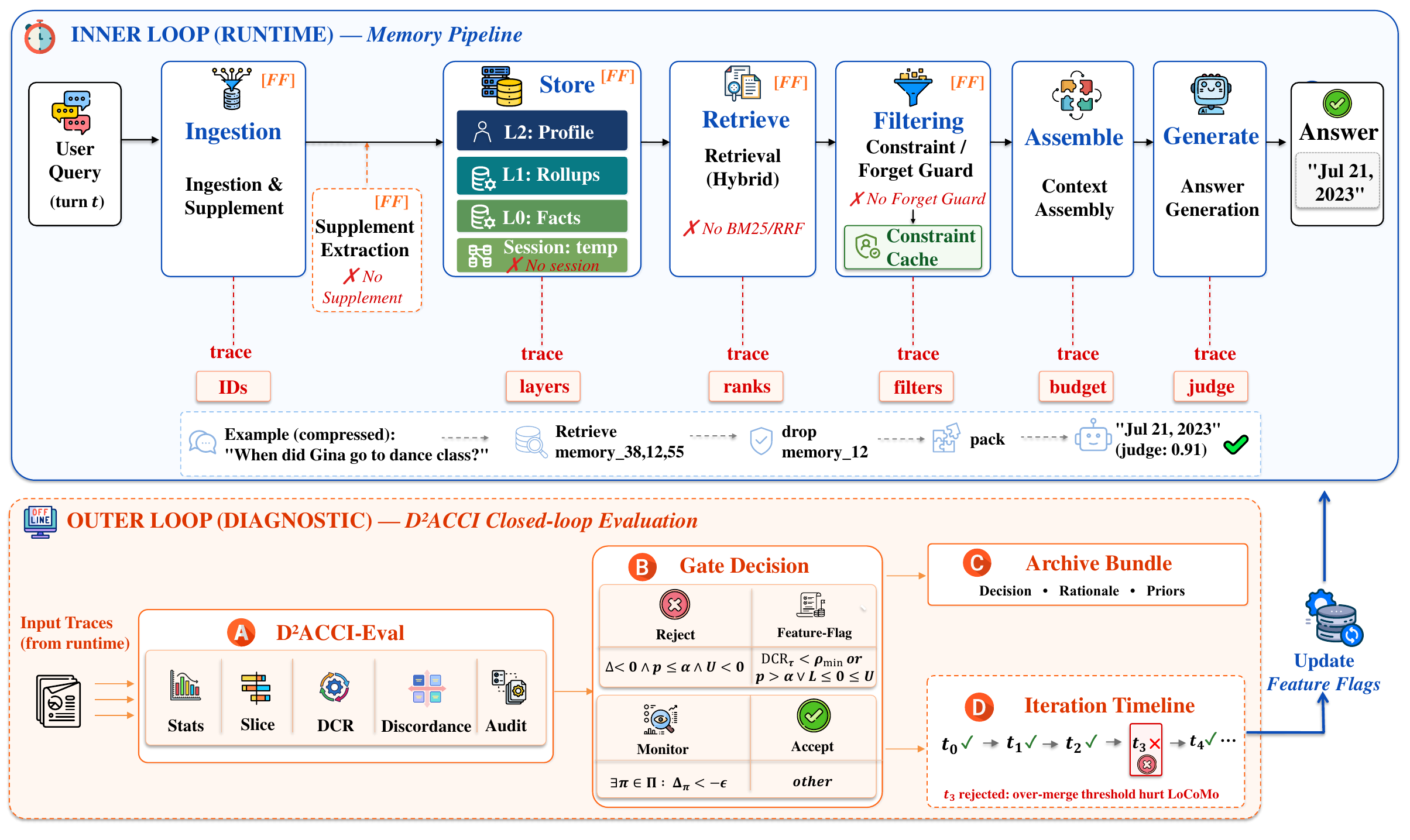}
\caption{\MiMemoryStack{} under the \DACCI{} dual-loop protocol. \textbf{Top}: the inner memory loop processes queries through feature-flagged stages (FF; independently togglable by the outer gate): ingestion, multi-layer storage, retrieval, constraint filtering, context assembly, and generation; each emits a typed diagnostic trace (IDs, layers, ranks, filters, budget, judge). Ablation targets (\ding{55}) mark paired ablations in Table~\ref{tab:ablations}. \textbf{Bottom}: the outer diagnostic loop feeds traces into \DACCI{}-Eval (paired stats, slice deltas, \DCov{}, discordance, audit), producing a gate decision that updates inner-loop feature flags for iteration $t{+}1$. The iteration timeline shows automatic gate replay including the $t_3$ rejection.}
\label{fig:architecture}
\end{figure*}

\section{D\textsuperscript{2}ACCI Framework}

\subsection{Problem Formulation}
Let $\mathcal{S}=\{s_i\}_{i=1}^{N}$ denote an evaluation set, where each sample $s_i=(q_i,h_i,y_i^*)$ contains a query, interaction history, and target answer.  
A memory configuration $\theta$ defines a pipeline $\mathcal{P}_{\theta}$ that produces $\hat{y}_i=\mathcal{P}_{\theta}(q_i,h_i)$. 
An evaluator $J$ yields a score $m_i=J(\hat{y}_i,y_i^*)$ and aggregate metric $\mathcal{M}(\theta)=N^{-1}\sum_i m_i$.

Aggregate performance alone does not reveal why a memory-augmented answer fails.
\DACCI{}  therefore records a diagnostic state for each sample:
\begin{equation}
 d_i=(e_i^{raw}, e_i^{store}, r_i^{top-k}, f_i^{filter}, c_i^{ctx}, g_i^{gen}, j_i),
\end{equation}
where $e_i^{raw}$ denotes evidence in the history, $e_i^{store}$ denotes stored memory, $r_i^{top-k}$ denotes retrieved items, $f_i^{filter}$ denotes filtering or reranking decisions, $c_i^{ctx}$ denotes assembled context, $g_i^{gen}$ denotes the generated answer, and $j_i$ denotes the final judgment.

A candidate $\theta'$ is compared against a baseline $\theta$ on paired samples, partitioning them into $I$ (improved), $R$ (regressed), $B_w$ (both wrong), and $B_c$ (both correct).  For each material regression or persistent error, the earliest actionable failure stage is labeled as ingestion miss, retrieval miss, context-assembly error, constraint error, generation error, component hurt, or unresolved ambiguity.  Algorithm~\ref{alg:gate_contract} first flags incomplete reports, then applies five ordered checks: reject significant harm, feature-flag low \DCov{}(Diagnostic Coverage Rate), feature-flag inconclusive statistics, accept-with-monitor protected-slice regressions, or accept otherwise.  The \DCov{} check precedes the inconclusive fallback, so positive but poorly observable candidates are still flagged.  Protected slices $\Pi$ are pre-specified from benchmark category structure (e.g., multi-hop, temporal, sensitive-preference) before any gate is evaluated.

\subsection{Dual-loop decomposition}
The key design choice is to separate two coupled but different loops.  The \emph{inner memory loop} runs at inference or ingestion time: it extracts candidate memories from user interactions, writes or updates multi-granularity stores, retrieves evidence for a query, applies update/forget constraints, assembles context, and generates an answer.  The \emph{outer diagnostic loop} runs over paired artifacts: it compares a feature-bearing run against an ablation, checks statistical and slice gates, inspects traces for the earliest evidence-loss stage, and then accepts, rolls back, or keeps a feature behind a flag.  

From an agent-architecture perspective, the outer loop is a \emph{metacognitive self-regulation layer}: the agent monitors its own competence boundaries, detects protected-slice regressions, and gates modifications to its own harness before they propagate.  This differs from a single metric-driven development loop because the outer loop can reject a component that raises a raw dashboard number but lacks a trace-supported mechanism or regresses a protected slice.  It also differs from pure prompt/program optimization: the outer loop does not search only over prompts, but controls architectural features such as supplement extraction, deduplication thresholds, retrieval channels, and forget-guard injection.  

Once instrumented, the gate can be triggered automatically after each candidate run, enabling automatic gate replay for candidate memory-policy changes with statistical regression controls.
Algorithm~\ref{alg:dacci-iteration} summarizes the complete dual-loop iteration from candidate construction to gated promotion, rollback, or feature-flagging.
Gate replay is deterministic under fixed thresholds and pre-specified protected slices.

\begin{algorithm}[t]
\caption{\DACCI{}-Eval gate-replay decision contract}
\label{alg:gate_contract}
\begin{algorithmic}[1]
\REQUIRE $\mathcal{R}=(\Delta,p,[L,U],\{\Delta_\pi\}_{\pi\in\Pi},\mathrm{DCR}_\tau)$; IDs, traces, $\Pi$, $\alpha,\rho_{\min},\epsilon$
\ENSURE Decision $D$, rationale $r$, archive $A$
\IF{$\mathcal{R}$ incomplete}
  \STATE $(D,r)\leftarrow(\textsc{Feature-Flag},\text{incomplete report})$
\ELSIF{$\Delta < 0 \wedge p \le \alpha \wedge U < 0$}
  \STATE $(D,r)\leftarrow(\textsc{Reject},\text{significant harm})$
\ELSIF{$\mathrm{DCR}_\tau < \rho_{\min}$}
  \STATE $(D,r)\leftarrow(\textsc{Feature-Flag},\text{insufficient trace})$
\ELSIF{$p > \alpha \vee L \le 0 \le U$}
  \STATE $(D,r)\leftarrow(\textsc{Feature-Flag},\text{inconclusive evidence})$
\ELSIF{$\exists \pi\in\Pi: \Delta_\pi < -\epsilon$}
  \STATE $(D,r)\leftarrow(\textsc{Accept-with-Monitor},\text{slice guard})$
\ELSE
  \STATE $(D,r)\leftarrow(\textsc{Accept},\text{positive paired evidence})$
\ENDIF
\STATE $A \leftarrow \mathrm{Archive}(\mathcal{R},\Pi,D,r,\text{checks, monitors, priors})$
\RETURN $D,r,A$
\end{algorithmic}
\end{algorithm}

\subsection{\MiMemoryStack{} System Instantiation}

\MiMemoryStack{} instantiates \DACCI{} as a shared memory kernel with thin benchmark adapters (Figure~\ref{fig:architecture}).  We retain only the design choices needed to interpret the ablations: L0 atomic facts, L1 topic rollups, L2 user profiles, and session memory are separate because each layer isolates a different failure mode.  No Supplement disables the long-session supplement stage and conservative deduplication; No Session disables answer-time session-memory retrieval; No Forget Guard disables constraint-cache filtering.

\paragraph{Design lessons.}
Three choices enable stage-level failure localization: (1)~\emph{layer separation}---each ablation maps onto a distinct failure family; (2)~\emph{feature flags}---retrieval channels, supplement extraction, and constraint injection are toggled per-run for cheap paired ablation; (3)~\emph{trace emission at every stage boundary}---without all three, a pipeline may improve accuracy while hiding whether the gain came from better evidence or lucky prompt formatting.

\paragraph{Diagnosable ingestion, retrieval, and constraints.}
L0 facts are extracted from raw turns with source-turn IDs; L1 rollups aggregate session-level patterns; L2 profiles distill stable preferences from L1.  Session memory is flushed at session boundaries as a structured record (raw turns, summary, keywords, timestamps, multi-granularity embeddings); long-session repair is handled by supplement extraction, which writes additional source-grounded L0 facts.  At answer time, vector/BM25/RRF/session retrieval produces candidates; constraint-cache filtering removes forgotten or superseded memories; context assembly packs constraints before ordinary evidence under a token budget.  Traces record source IDs, memory IDs, top-$k$ scores, filtered items, packed/dropped context, and the final prompt at each stage boundary; these are exactly the fields consumed by \DCov{} and gate replay.

\subsection{\DACCI{}-Eval and Diagnostic Coverage}
Ordinary A/B testing answers whether a candidate is better on average; it does not say whether the pipeline preserved enough evidence to debug the errors it still makes.  \DACCI{}-Eval is the artifact layer for this distinction.  Given paired details files, it joins examples by stable IDs, computes aggregate and slice-level deltas, exports Full-only and Ablated-only discordances, runs McNemar tests and bootstrap confidence intervals, and creates a root-cause audit queue.  Its additional observability metric is Diagnostic Coverage Rate (\DCov{}).  For a failure or discordance set $F$, stage set $S$, trace $d_i$, and a per-stage actionability predicate $A(d_i,s)$, let $C(d_i)=\sum_{s\in S}\mathbb{1}[A(d_i,s)]$ be the number of actionable stages.  We define
\begin{equation}
\mathrm{DCR}_\tau(F;S)=\frac{1}{|F|}\sum_{i\in F}\mathbb{1}\left[C(d_i)\ge \tau\right].
\end{equation}
Here $A$ is true when the trace contains the minimum fields needed to inspect a stage: source-turn or memory IDs for ingestion, top-$k$ items and scores for retrieval, filter decisions for reranking, final packed context for assembly, constraint IDs for forget/update guards, and output plus judge metadata for generation.  In the artifact, $A(d_i,s)$ is schema-validated rather than free-form: a stage is actionable only when the required IDs, scores, decisions, and metadata are present.  \DCov{} therefore measures observability rather than accuracy.  A high-accuracy system with low \DCov{} may be hard to repair; a lower-accuracy but high-\DCov{} run can still be valuable if it localizes missing evidence.  Unless otherwise stated, gate replay uses \DCov{}@3, requiring at least three actionable stages, and a fixed minimum coverage threshold of 0.90 set before inspecting paired results; sensitivity is reported at \DCov{}@1, @5, and @6.  To make the number interpretable, \DACCI{}-Eval includes a results-only contrast that retains the answer and score but removes stage traces; this contrast receives zero coverage under the default three-stage predicate.

\paragraph{Design properties of \DCov{}.}
Three properties distinguish \DCov{} from an ad-hoc coverage count and justify its use as a gate criterion:
(1)~\emph{Stage monotonicity}---adding instrumentation stages can only increase $C(d_i)$, so \DCov{} never penalizes a more observable system ($\mathrm{DCR}_\tau(F;S')\ge\mathrm{DCR}_\tau(F;S)$ for $S'\supset S$).
(2)~\emph{Accuracy-orthogonal separation}---\DCov{} captures information that accuracy cannot: the Mem0-compatible slice achieves 43.33\% accuracy under both trace-adapted and result-only configurations, yet \DCov{}@3 is 100\% vs.\ 0\%.
(3)~\emph{Gate soundness}---under Algorithm~\ref{alg:gate_contract}, a null component (true $\Delta{=}0$) is promoted with probability at most $\alpha$, because promotion requires McNemar $p\le\alpha$ with CI excluding zero; the false-promotion rate is bounded by the test's Type-I control.
Together, these ensure \DCov{} is well-behaved and non-redundant, and that the outer-loop gate provides a statistical guarantee against adopting inert components---distinguishing it from metric-only dashboard decisions.




\begin{table*}[t]
\centering

\scriptsize
\setlength{\tabcolsep}{5pt}
\begin{tabular}{@{}lccccccccc@{}}
\toprule
Benchmark & Eval set & Ours & MemBrain & $\Delta$ & \multicolumn{4}{c@{}}{Representative slices (category / (\textcolor{blue}{count}, \textcolor{red}{accuracy}))} \\
\midrule
\multirow{2}{*}{\locomo{}} & \multirow{2}{*}{1540 non-adv QA} & \multirow{2}{*}{\textbf{93.59\%}} & \multirow{2}{*}{93.25\%} & \multirow{2}{*}{+0.34pp} &
  \cellcolor{gray!10}Single-hop & \cellcolor{gray!20}Multi-hop & \cellcolor{gray!10}Temporal & \cellcolor{gray!20}Open \\
 & & & & &
  \cellcolor{gray!10}(\textcolor{blue}{841}, \textcolor{red}{96.2}) & \cellcolor{gray!20}(\textcolor{blue}{282}, \textcolor{red}{89.5}) & \cellcolor{gray!10}(\textcolor{blue}{321}, \textcolor{red}{92.4}) & \cellcolor{gray!20}(\textcolor{blue}{96}, \textcolor{red}{86.8}) \\
\arrayrulecolor{gray!30}\hline\arrayrulecolor{black}
\multirow{2}{*}{\lme{}} & \multirow{2}{*}{500/500 questions} & \multirow{2}{*}{\textbf{90.93\%}} & \multirow{2}{*}{85.60\%} & \multirow{2}{*}{+5.33pp} &
  \cellcolor{gray!10}Single-user & \cellcolor{gray!20}Knowledge-update & \cellcolor{gray!10}Multi-session & \cellcolor{gray!20}Temporal \\
 & & & & &
  \cellcolor{gray!10}(\textcolor{blue}{70}, \textcolor{red}{99.5}) & \cellcolor{gray!20}(\textcolor{blue}{78}, \textcolor{red}{93.6}) & \cellcolor{gray!10}(\textcolor{blue}{133}, \textcolor{red}{85.0}) & \cellcolor{gray!20}(\textcolor{blue}{133}, \textcolor{red}{89.0}) \\
\arrayrulecolor{gray!30}\hline\arrayrulecolor{black}
\multirow{2}{*}{\pmem{}} & \multirow{2}{*}{200 personas, 5000 MCQ} & \multirow{2}{*}{\textbf{57.20\%}} & \multirow{2}{*}{55.72\%} & \multirow{2}{*}{+1.48pp} &
  \cellcolor{gray!10}Sensitive & \cellcolor{gray!20}Forget & \cellcolor{gray!10}Neutral & \cellcolor{gray!20}Therapy \\
 & & & & &
  \cellcolor{gray!10}(\textcolor{blue}{511}, \textcolor{red}{86.3}) & \cellcolor{gray!20}(\textcolor{blue}{1048}, \textcolor{red}{66.9}) & \cellcolor{gray!10}(\textcolor{blue}{858}, \textcolor{red}{51.2}) & \cellcolor{gray!20}(\textcolor{blue}{627}, \textcolor{red}{44.5}) \\
\bottomrule
\end{tabular}
\caption{Main results and representative slice breakdowns. External rows are reference points, not fully normalized same-stack baselines. Slice-level reporting is required because memory interventions are frequently non-monotonic.}
\label{tab:main-results}
\end{table*}

\section{Experiments}
\subsection{Experimental Setup}
We evaluate the same shared core across three public regimes.  \locomo{} tests long-term conversational QA over multi-session dialogues \citep{maharana2024locomo}.  We report non-adversarial judge accuracy on 1540 questions from 10 conversations.  \lme{} evaluates chat assistants on long-term interactive memory and contains 500 curated questions in the LongMemEval-S setting \citep{wu2025longmemeval}.  \pmem{} evaluates implicit persona memory with 5000 multiple-choice questions over 200 personas \citep{jiang2025personamem}.

Unless otherwise stated, the answer model is GPT-4.1-mini, the open-ended judge is GPT-4o-mini with three deterministic runs per question (majority vote), and embeddings use BGE-M3 \citep{chen2024bge}.  \pmem{} uses exact-match multiple-choice accuracy.  Paired confidence intervals use BCa bootstrap with 10{,}000 resamples; McNemar tests use the exact two-sided binomial form.  External rows are reference points from published papers or  public summaries; our strongest claims rely on internal paired comparisons using the same artifacts, prompts, and model settings.

MemBrain is our primary public reference point on all three benchmarks; its numbers are taken from its maintained evaluation summary under the same splits~\citep{feelingai2026membrain}. 
We treat published external numbers as reference context, not controlled baselines; the controlled evidence comes exclusively from internal paired ablations where the only variable is one feature flag.
Concurrent systems (Mem0 Platform 92.5\%/94.4\% on \locomo{}/\lme{}~\citep{chhikara2025mem0}; EverMemOS 92.3\%/83.0\%~\citep{evermemos2026acl}) are not directly comparable due to differences in answer models, judge prompts, and retrieval budgets.  We retain MemBrain as the primary reference point because it documents conditions closest to ours across all three benchmarks.

Each run produces per-question JSONL and summary metadata; \DACCI{}-Eval consumes to produce paired statistics, slice deltas, discordance lists, \DCov{} reports, and audit templates.  We report results in terms of these artifacts below.


\begin{table}[t]
\centering

\scriptsize
\setlength{\tabcolsep}{3pt}
\begin{tabular}{lcc}
\toprule
Condition vs. Human A & Fine $\kappa$ & Coarse $\kappa$ \\
\midrule
Enriched GPT-4o trace relabel & 0.571 & 0.619 \\
Result-only GPT-4o relabel & 0.258 & 0.272 \\
Human B & 0.674 & 0.711 \\
\bottomrule
\end{tabular}
\caption{Audit-label contrast on the 60-case packet.  Human~B is an independent annotator; all rows are pre-adjudication.}
\label{tab:audit-contrast-main}
\end{table}

\subsection{Main Results and Paired Ablations}
Table~\ref{tab:main-results} summarizes the evaluated configuration and representative slice breakdowns.  The \locomo{} margin over the reference point is within noise (+0.34pp) and is not claimed as a contribution; the paper's value comes from the ablation, diagnostic-coverage, and audit analyses below.  The system is strongest on single-hop \locomo{} and single-session-user \lme{} questions, but weaker on open-domain \locomo{}, multi-session \lme{}, and therapy/background \pmem{} cases.  These weaknesses define protected-slice monitors and motivate the root-cause audit.

Table~\ref{tab:ablations} reports completed paired ablations only.  The sign convention is Full minus Ablated, so a positive delta means the evaluated component helps.  
Negative/null results are intentionally retained---they are central to \DACCI{}: a component should not be promoted unless paired evidence supports its mechanism.
The No-Session replay disables session-memory retrieval and session-context injection at answer time while retaining all stored memories, BM25/RRF, agentic retrieval, model settings, and judge settings; it isolates whether cross-session state contributes to multi-session and temporal questions without a re-ingestion confound.


\begin{table*}[t]
\centering
\begin{threeparttable}

\scriptsize
\setlength{\tabcolsep}{3pt}
\renewcommand{\arraystretch}{1.15}

\begin{tabular*}{\textwidth}{
    @{\extracolsep{\fill}}
    ll
    r
    c
    c
    c
    c
    c
    l
    l
    @{}
}
\toprule
\multirow{2}{*}{Benchmark}
& \multirow{2}{*}{Comparison}
& \multirow{2}{*}{$n$}
& \multirow{2}{*}{Ablated}
& \multirow{2}{*}{$\Delta$}
& \multicolumn{3}{c}{Paired evidence}
& \multicolumn{2}{c}{Gate decision} \\
\cmidrule(lr){6-8}
\cmidrule(l){9-10}
&
&
&
&
&
$p$
& 95\% CI
& Full wins / Ablated wins
& Metric-only
& \DACCI{} gate \\
\midrule

LoCoMo
& No BM25/RRF
& 1,540
& 94.00\%
& $-0.41$ pp
& .4426
& [$-1.36$, 0.56]
& 27/34
& Demote
& Feature-flag \\

LoCoMo
& No Supplement
& 1,540
& 90.89\%
& $+2.71$ pp
& .0009
& [1.26, 4.20]
& 90/50
& Promote
& Accept; MH monitor \\

LME
& No BM25/RRF
& 500
& 89.93\%
& $+1.00$ pp
& .4583
& [$-1.13$, 3.13]
& 17/12
& Promote
& Feature-flag \\

LME
& No Session
& 500
& 87.27\%
& $+3.67$ pp
& .0026
& [1.33, 6.00]
& 28/9
& Promote
& Accept; KU monitor \\

PMem
& No Forget Guard
& 5,000
& 55.28\%
& $+1.92$ pp
& .0030
& [0.64, 3.18]
& 560/464
& Promote
& Accept; slice monitor \\

\bottomrule
\end{tabular*}


\end{threeparttable}
\caption{Completed paired diagnostic ablations with metric-only vs.\ \DACCI{} gate decisions.  CI is bootstrap 95\% for Full$-$Ablated; Full wins / Ablated wins count correct examples.  Metric-only: accept if $\Delta{>}0$, demote if $\Delta{<}0$; \DACCI{} adds statistical and slice-level checks.}
\label{tab:ablations}
\end{table*}

The completed ablations change the interpretation of the system.  Each ablation targets the benchmark whose category structure most directly stresses the removed component: supplement extraction addresses long-session evidence gaps central to \locomo{}'s multi-hop questions; session retrieval addresses cross-session temporal reasoning in \lme{}; Forget Guard addresses preference revision central to \pmem{}.  BM25/RRF is tested on both open-ended benchmarks and is null on both, retained only as a monitored feature flag.  The three accepted components are validated with slice monitoring for the small non-monotonic regressions in Table~\ref{tab:ablations}; cross-benchmark replication (e.g., No Session on \locomo{}) is supported identically by the protocol but not yet complete.  
The audit-label contrast in Table~\ref{tab:audit-contrast-main} provides the key validation; supplementary material includes per-queue details and a separate 110-example \locomo{} run. This supports our central claim: useful memory changes require paired, slice-level diagnostics, not aggregate wins alone.

BH correction over the five tests preserves the accepted rows at $q{=}0.01$ (adjusted $p\le.0050$) and leaves BM25/RRF null; we therefore read BM25/RRF as not promotable, not as zero-effect proof.  An alternate GPT-4.1-mini replay over all 267 original open-ended discordances preserves the four open-ended gate outcomes (BM25/RRF feature-flag $p=.175/.115$; Supplement/Session accept $p=.0014/.0009$), with concordant examples left unrescored.

\subsection{Diagnostic Coverage and Gate Validation}

\paragraph{Trace availability and root cause agreement.}
Table~\ref{tab:audit-contrast-main} compares root-cause labeling under two conditions on the 60-case packet: full enriched traces (retrieval results, assembly context, constraint records) versus result-only information (answer and score).  Human--human pre-adjudication agreement across the 60 labeled cases is 44/60 exact matches ($\kappa{=}0.674$ fine, $0.711$ coarse; bootstrap 95\% CIs $[0.534,0.797]$ and $[0.564,0.841]$).  The trace effect remains large: enriched-trace GPT-4o relabeling reaches $\kappa{=}0.571$ fine ($0.619$ coarse; CIs $[0.418,0.715]$ and $[0.464,0.764]$) versus $\kappa{=}0.258$ ($0.272$; CIs $[0.159,0.355]$ and $[0.175,0.376]$) for result-only.  Under the result-only GPT-4o relabeling condition, 36/60 cases are assigned to ``unresolved'', because answer-level evidence alone does not identify failing stage.  This validates the core premise: trace availability transforms root-cause labeling from guesswork into an auditable task.

The following \locomo{} discordance illustrates that result-only outputs show only a date flip, whereas traces reveal that supplement extraction preserved the decisive temporal evidence.
At the aggregate level, the gate accepts Supplement with a multi-hop monitor over 1540 pairs (+2.71pp, $p{=}.0009$, \DCov{}@3=99.47\%).




\newtcolorbox{examplebox}{
    colback=blue!3!white,  
    colframe=blue!70!black, 
    fonttitle=\bfseries,
    title=Case Study (LoCoMo conv1:q038),
    boxrule=1pt,
    top=3pt,
    left=6pt,
    right=6pt,
    bottom=3pt,
    before skip=5pt,     
    after skip=5pt,       
    boxsep=3pt             
}

\newcommand{\question}[1]{\textcolor{blue}{\textbf{Question:}} #1\par\vspace{-3pt}}
\newcommand{\goldresponse}[1]{\textcolor{green!60!black}{\textbf{Gold Response:}} #1  \par\vspace{-3pt}}
\newcommand{\ablatedresponse}[1]{\textcolor{red!70!black}{\textbf{Ablated  Response(No Supplement):}} #1 \textcolor{red}{$\times$} \par\vspace{-3pt}}
\newcommand{\fullresponse}[1]{\textcolor{cyan!60!black}{\textbf{Full Response(Supplement enabled):}} #1 \textcolor{green!60!black}{$\checkmark$} \par\vspace{-3pt}}
\newcommand{\diagnosis}[1]{\textcolor{purple!70!black}{\textbf{\DACCI{} Diagnosis:}} #1}

\begin{examplebox}
    \question{When did Gina go to a dance class with friends?}\par\medskip

    \goldresponse{21 Jul 2023.}\par\medskip
    
    \ablatedresponse{Jul 14. Final context carries a wrong 2023-07-14 temporal memory.}\par\medskip
    
    \fullresponse{Jul 21. Trace keeps source-backed 2023-07-21 evidence. }\par\medskip
    \diagnosis{Result-only outputs show a date flip but not its cause; traces show the supplement stage preserves the decisive temporal evidence.}\par\medskip

\end{examplebox}

\paragraph{Diagnostic coverage and trace masking.}
The audit contrast above establishes that trace availability materially improves diagnostic consistency in the 60-case packet ($\kappa$ jumps from 0.258 to 0.571).  \DCov{} formalizes this as a measurable proxy: it tracks whether each failure retains enough stage evidence for inspection, not whether the trace itself is a correct root-cause label.  Crucially, \DCov{} is \emph{graded}, not binary.  
Counterfactual trace masking produces monotonic degradation across the five paired ablations without requiring additional model runs. Result-only records achieve 100.0\% at \DCov{}@1 but 0.0\% at @3, @5, and @6. Retrieval-only traces achieve 98.4--100.0\% at @1 and @3 but 0.0\% at @5 and @6. Adding stored-memory, retrieval, and assembled-context traces extends the 98.4--100.0\% coverage through @5, whereas only full traces maintain 98.4--100.0\% coverage at @6. The 98.4\% vs.\ 100\% variation across ablations is the most informative signal---it identifies samples that bypass a given stage (e.g., short conversations that never trigger supplement extraction) and are therefore only partially diagnosable.  These partial-observability blind spots would be invisible under a binary ``logged vs.\ not-logged'' check; \DCov{} surfaces them as concrete instrumentation targets for the next iteration.  Pair-level \DCov{}@3 is 99.21\% over 127 \locomo{} No-BM25/RRF discordances, 99.47\% over 190 \locomo{} No-Supplement cases, 98.63\% over 73 \lme{} No-Session cases, and 100.00\% over 2700 \pmem{} No-Forget-Guard cases; the separate 110-example \locomo{} result-only check is 0/110, confirming that scalar outcomes alone provide no localizable evidence.


Beyond diagnostic coverage, we validate that the gate itself produces meaningful decisions.  The last two columns of Table~\ref{tab:ablations} contrast the dual-loop gate against a metric-only policy.  The metric-only policy issues contradictory BM25/RRF verdicts across benchmarks (demote on \locomo{}, promote on \lme{}); the dual-loop gate identifies both as null and retains a monitored feature-flag.  For the three accepted components, metric-only promotes unconditionally, while the dual-loop gate attaches explicit slice monitors (MH $-2.6$pp, KU $-1.3$pp, multi-slice non-monotonicity in \pmem{}) that prevent silent regressions from propagating.  A threshold sweep confirms that the accept/feature-flag split is stable for $\alpha\ge0.005$; only monitoring annotations change as $\epsilon$ varies.

\subsection{Trace-Guided Failure Analysis and Iteration}
The preceding sections validate the gate statistically; we now show how traces guide specific failure localization and repair.

\paragraph{Failure boundary analysis.}
A rule-based trace pass over 2,140 wrong \pmem{} cases localizes the performance boundary: generation/option-scoring accounts for 1,223 cases (57.1\%), constraint handling for 637 (29.8\%), retrieval misses for 278 (13.0\%), and ingestion misses for 2 (0.1\%).  \pmem{} is included precisely because it stress-tests the protocol under conditions where memory-only improvements hit diminishing returns.  This diagnosis is itself a protocol output: without per-question retrieval and assembly traces, one cannot distinguish ``never stored'' from ``stored but model picked the wrong option.''  The practical payoff is preventing wasted optimization: \DACCI{} localizes the boundary between memory-addressable and generation-addressable errors (87\% lie beyond retrieval), a conclusion invisible under aggregate-only evaluation and one that redirects engineering effort toward the actual bottleneck.
\paragraph{Ingestion stage over-merge.}
Traces for affected questions (e.g., ``What breed is James's second dog?'') revealed that L0 contained a single merged memory ``James has dogs'' instead of three separate records.  Retrieval returned this merged record (rank 1, score 0.71) but lacked breed information.  Failure localized to \textbf{ingestion-stage over-merge}; the repair (threshold 0.92) recovered 4.48pp.

\paragraph{Constraint stage filtering.}
In the PersonaMem follow-up audit (e.g., \texttt{persona136:q7}), the full trace's forget scan preserves the Diwali option and the model selects the gold answer, while the no-guard variant drifts to a different cultural-event option.  This localizes the gain to the \textbf{constraint} stage rather than retrieval: the evidence was already present, but the guard prevented a forgotten-topic spillover.

\begin{figure}[htbp]
    \centering
    \includegraphics[width=0.45\textwidth]{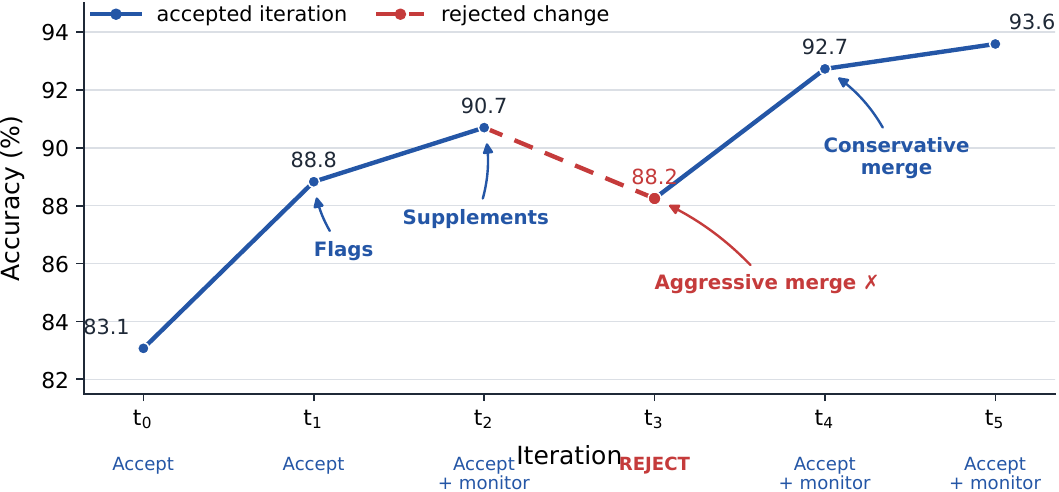}
    \caption{\locomo{} iteration trajectory with automatic gate-replay decisions.  Each transition is produced by the replay script ($<$1\,s per candidate); all five decisions match the human-in-the-loop outcome.  The dashed red segment marks $t_3$: traces localized over-merging, triggering automatic rejection.}
\label{fig:trajectory}
\end{figure}

\begin{table}[t]
\centering

\scriptsize
\setlength{\tabcolsep}{5pt}
\begin{tabularx}{\linewidth}{@{} 
    l                          
    >{\hsize=0.7\hsize}X       
    >{\hsize=1.3\hsize}X       
    l                          
@{}}
\toprule
\textbf{AP$^{*}$} & \textbf{Trigger} & \textbf{Observed Behavior} & \textbf{Resolution} \\
\midrule
\multirow{2}{*}{OM} & \multirow{2}{*}{Threshold $(0.75)$} & Facts collapse into a single memory representation & \multirow{2}{*}{\textbf{Reverted}} \\
\midrule
DA & Option-aware filter & Retrieval surfaces distractor evidence & \textbf{Rejected} \\
\midrule
\multirow{2}{*}{NME} & \multirow{2}{*}{Gated expansion} & Recovers some targets, introduces off-target noise ($57.34\% \to 57.24\%$) & \multirow{2}{*}{\textbf{Slice-gated}} \\
\bottomrule
\multicolumn{4}{l}{\footnotesize $^{*}$ AP: Anti-pattern} \\
\end{tabularx}
\caption{Representative archived anti-patterns (OM: Over-merge, DA: Distractor amplification, NME: Non-monotonic expansion) exposed by \DACCI{} traces.}
\label{tab:negative}
\end{table}




Figure~\ref{fig:trajectory} shows the \locomo{} trajectory; the $t_3$ rejection (detailed in the case study above) is the trace-guided outer-loop example.  The No-Supplement row in Table~\ref{tab:ablations} is its paired statistical counterpart ($+2.71$pp, $p{=}.0009$).

To verify the outer loop requires no human intervention, we replay the gate-decision script over \locomo{} iterations ($t_0$--$t_5$) using only paired reports (deltas, $p$-values, slice vectors, \DCov{}) as input.  
The script produces decisions matching human judgment in $<$1\,s.  Figure~\ref{fig:trajectory} annotates each transition with its automatic gate-replay outcome; all five decisions match the human-in-the-loop process, including the critical $t_3$ rejection.  Once instrumented, the gate script reproduces human decisions without per-candidate manual review.
Table~\ref{tab:negative} catalogs rejected interventions as reusable anti-patterns.

\section{Related Work}
\paragraph{Agent memory benchmarks and systems.}
Long-term memory benchmarks evaluate agents' ability to retain, retrieve, and update information across extended interactions~\citep{maharana2024locomo,wu2025longmemeval,jiang2025personamem}.
Corresponding systems explore virtual context hierarchies, add/search/update APIs, dynamically linked notes, temporal or graph stores, and multi-granular representations~\citep{packer2023memgpt,zhong2023memorybankenhancinglargelanguage,rasmussen2025zeptemporalknowledgegraph,chhikara2025mem0,xu2025amemagenticmemoryllm,kang2025memoryosaiagent,hu2026evermemosselforganizingmemoryoperating,wang2026memmachine,xu2026napmem,xia2026memora}.  
RL-based work further optimizes memory management, utility estimation, retrieval, answer generation, and proactive querying~\citep{zhang2026memrlselfevolvingagentsruntime,yan2026memoryr1enhancinglargelanguage,zhang2026learningrememberendtoendtraining,yu2026agenticmemorylearningunified,wang2026explorelongtermmemorybenchmark}.
These systems make memory decisions adaptive, but task-level rewards and aggregate scores can conflate ingestion, retrieval, constraint, and generation failures; \DACCI{} is complementary because it supplies paired, slice-level, trace-preserving acceptance criteria.

\paragraph{Self-evolving memory architectures and LM programs.}
Recent agents evolve memories, reusable skills, or the memory mechanism itself from interaction feedback, reflective search, accumulated experience, and failure signals~\citep{shinn2023reflexionlanguageagentsverbal,wang2023voyageropenendedembodiedagent,zhao2024expelllmagentsexperiential,wei2026evomemorybenchmarkingllmagent,zhang2026memskilllearningevolvingmemory,cheng2026mem2evolveselfevolvingagentscoevolutionary,xiong2026learningcontinuallylearnmetalearning,zhang2025memevolvemetaevolutionagentmemory,pan2026mstartaskdeservesmemory,liu2026evolvememselfevolvingmemoryarchitectureautoresearch}.
Similarly, DSPy, MIPRO, and TextGrad optimize language-model programs or instructions under task objectives~\citep{khattab2024dspy,opsahlong2024mipro,yuksekgonul2024textgrad}. 
\DACCI{} differs by making paired statistical testing, protected-slice monitoring, null-result preservation, and trace coverage explicit promotion criteria rather than prescribing a search space.


\paragraph{Observability and regression-controlled evolution.}
Observability tools and self-evolving-agent research highlight the need for reliable evaluation and capability preservation~\citep{gao2025selfevolving,yu2026selfevolvingagentsforgetcapability}. 
MemTrace performs post-hoc attribution of failures to memory operations~\citep{deng2026memtrace}.
\DACCI{} complements such tools by combining stage-level traces with paired gates, protected-slice checks, archived null results, and \DCov{} to test whether a modification is reproducible, non-regressive, and diagnostically auditable.


\section{Conclusion}

We introduce \DACCI{} in this paper, a dual-loop protocol that turns memory-system changes into auditable decisions grounded in paired statistics, slice monitors, and stage traces.
Across three public agent memory benchmarks, the protocol distinguishes statistically validated improvements from null changes.
When instantiated in \MiMemoryStack{}, \DACCI{} achieves 93.59\% accuracy on \locomo{}, 90.93\% on \lme{}, and 57.20\% on \pmem{}.
The three accepted components yield statistically significant gains of 1.92–3.67 percentage points, while the corresponding diagnostic artifacts achieve 98–100\% \DCov{}@3, preserving sufficient stage-level evidence to localize failures and support trace-guided repair rather than blind tuning.
The protocol is portable to any memory stack that emits stable sample IDs and comparable stage traces.
In the future, the protocol will be extended to online settings and cross-stack comparisons.
In addition, we will develop automated trace verification to complement manual audits.

{\small\bibliography{references}}
\end{document}